\documentclass{article}

\usepackage[preprint,nonatbib]{neurips_2020}

\usepackage[utf8]{inputenc} 
\usepackage[T1]{fontenc}    
\usepackage{hyperref}       
\usepackage{url}            
\usepackage{booktabs}       
\usepackage{amsfonts}       
\usepackage{nicefrac}       
\usepackage{microtype}      

\usepackage{graphicx}
\usepackage{subfigure}
\usepackage{graphicx} 
\usepackage{amsmath}
\usepackage{bm}
\usepackage{multirow}
\usepackage{amssymb}
\usepackage{wrapfig}

\title{Distant Transfer Learning via Deep Random Walk}

\author{%
  Qiao Xiao and Yu Zhang\thanks{Corresponding author.}\\
  Department of Computer Science and Engineering \\
  Southern University of Science and Techonlogy \\
  \texttt{xiaoq3@mail.sustech.edu.cn},  \texttt{yu.zhang.ust@gmail.com} \\
}

\begin{document}

\maketitle

\begin{abstract}
Transfer learning, which is to improve the learning performance in the target domain by leveraging useful knowledge from the source domain, often requires that those two domains are very close, which limits its application scope. Recently, distant transfer learning has been studied to transfer knowledge between two distant or even totally unrelated domains via auxiliary domains that are usually unlabeled as a bridge in the spirit of human transitive inference that it is possible to connect two completely unrelated concepts together through gradual knowledge transfer. In this paper, we study distant transfer learning by proposing a DeEp Random Walk basEd distaNt Transfer (DERWENT) method. Different from existing distant transfer learning models that  implicitly identify the path of knowledge transfer between the source and target instances through auxiliary instances, the proposed DERWENT model can explicitly learn such paths via the deep random walk technique. Specifically, based on sequences identified by the random walk technique on a data graph where source and target data have no direct edges, the proposed DERWENT model enforces adjacent data points in a squence to be similar, makes the ending data point be represented by other data points in the same sequence, and considers weighted training losses of source data. Empirical studies on several benchmark datasets demonstrate that the proposed DERWENT algorithm yields the state-of-the-art performance.


\end{abstract}

\section{Introduction}




Transfer learning \cite{py10,yzdp20} aims to effectively enhance the performance of the target domain by learning useful knowledge from the source domain and it has a wide range of applications \cite{zllj19,ud10,plwy11}, especially when the target domain has limited label annotations. Using a large number of labeled data in the source domain to improve the performance in the target domain with limited or even on labeled training data via transfer learning models can greatly reduce the cost of labeling in the target domain.

A major assumption of traditional transfer learning is that the source and target domains should be close or similar to each other. When there is a large discrepancy between the target domain and the source domain, traditional transfer learning methods likely fail to work and even lead to the `negative transfer' phenomenon \cite{py10,yzdp20}. Instead, distant transfer learning \cite{tszy15,tzpy17} is proposed to handle this situation. Inspired by the transitive learning ability of human that two unrelated concepts can be connected via some intermediate concepts as a bridge, distant transfer learning uses data in auxiliary domains as such bridge to connect two distant domains, which makes the knowledge transfer between two distant domains possible. Distant transfer learning broadens the application scope of transfer learning and makes the learning system close towards human intelligence.


As pioneered in distant transfer learning, Tan et al. \cite{tszy15} require that auxiliary domain that includes both characteristics of target domain and source domain in a form of the co-occurrence data and propose a matrix-factorization-based model to achieve one-step transitive learning through the auxiliary domain. By relaxing such requirement on the data form in the auxiliary domain, the Distant Domain Transfer Learning (DDTL) method \cite{tzpy17} utilizes the idea of self-paced learning \cite{kpk10} to select both useful source and auxiliary data based on the reconstruction error to help improve the performance in the target domain which has limited labeled data. However, those two studies cannot explicitly identify the transfer path between the source and target domains via auxiliary domains.

\begin{figure}
    \centering
    \includegraphics[scale=0.35]{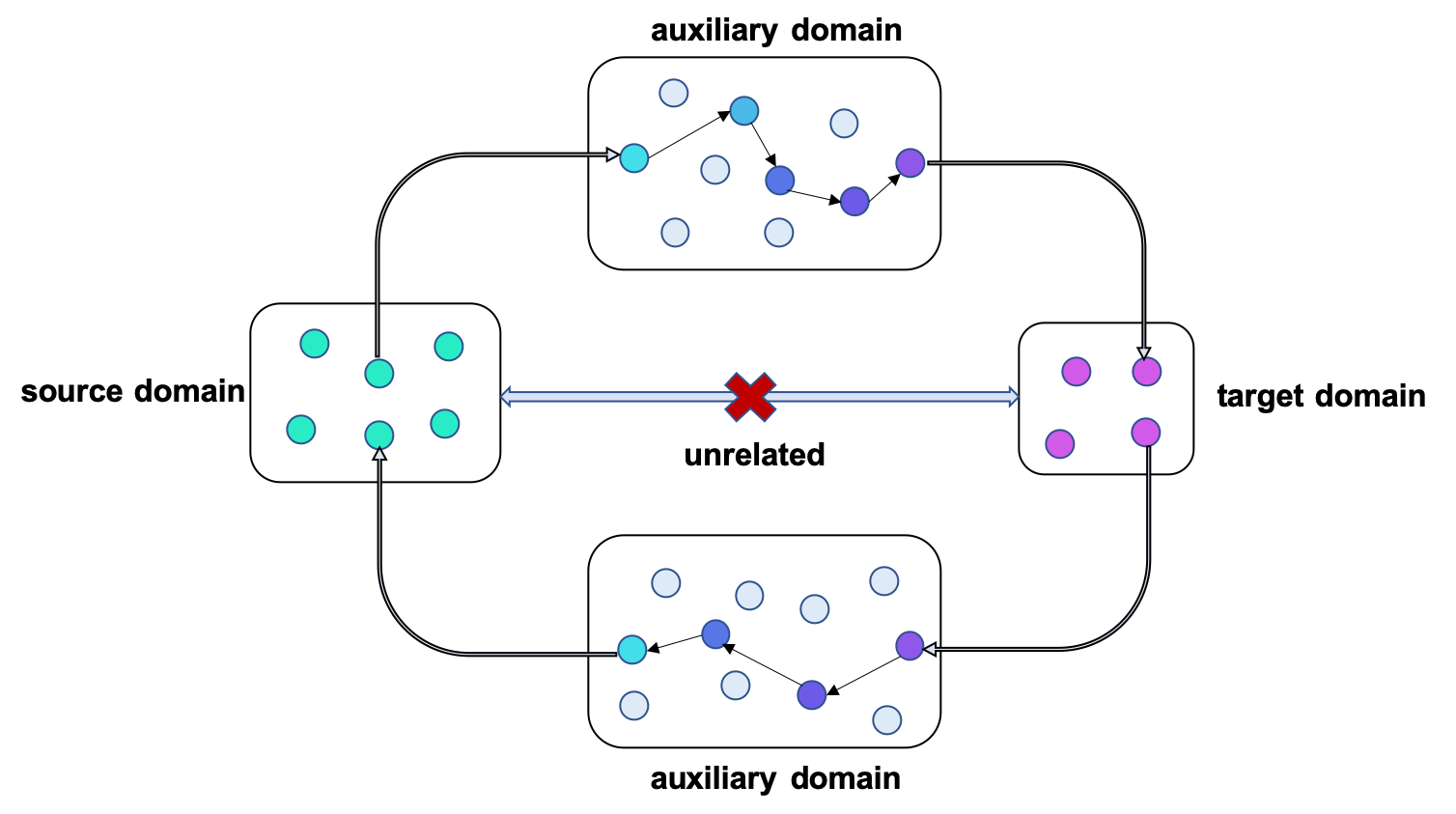}
    \vskip -0.1in
    \caption{An illustration of distant transfer learning. In distant transfer learning, the source and target domains cannot be directly transferred since the discrepancy between them is too large, making directly transfer fail to work. The proposed DERWENT method can automatically find the transferring path by deep random walk between the source domain and the target domain through auxiliary domains.}
    \label{fig:label1}
    \vskip -0.1in
\end{figure}

In this paper, we follow the setting of \cite{tzpy17} to study distant transfer learning with an objective to identify the transfer path between two distant domains, which is what previous studies cannot do. Identifying the transfer path can also improve the interpretability of the model by visualizing the transfer process. To achieve that, We adopt deep random walk to generate transfer paths between those two domains to complete the transfer learning. Specifically, as shown in the Figure \ref{fig:label1}, we construct a graph on all the data from all the domains with edge weights measuring similarities of pairs of data based on the hidden feature representation learned from a neural network. Note that there is no edge between source and target data in the graph as directly transferring is not so feasible. 
Then based on the constructed graph, we can generate sequences to connect source and target data through auxiliary data. For each sequence, the DERWENT model enforces adjacent data points in this sequence to be similar and make the ending data point in this sequence be represented by other data points via a Long Short-Term Memory (LSTM) \cite{hs97}. Moreover, the DERWENT model considers to incorporate the classification loss of source data in a squence with a weight depending on the data similarity.

In summary, the main contributions of this paper are three-fold.\\
1. We propose a novel DERWENT model for distant transfer learning by combining with deep random walk to generate transfer sequences between the source and target domains.\\ 
2. We conduct extensive experiments on twenty distant transfer learning tasks constructed from several benchmark datasets to validate the effectiveness of the proposed DERWENT model.\\
3. The proposed DERWENT model can identify the transfer path, which can improve the model interpretability by visualizing such sequences.

\section{Related Works}

 In transfer learning, given the source domain $\mathcal{D}_s$ and the target domain $\mathcal{D}_t$, it is usually assumed that their probability distributions are different. 
 There are mainly three typical types of transfer learning algorithms, including instance-based transfer learning \cite{dyxy07,kh16}, feature-based transfer learning \cite{pwty09,hy11}, and parameter-based transfer learning \cite{psyt08}. 
 Recently, transfer learning is mostly combined with deep neural networks \cite{thds15,lwcsy16}. However, the aforementioned work study traditional transfer learning which requires that the source and target domain are close and they may not achieve good performance under distant transfer learning.  

For distant transfer learning, Tan et al. \cite{tszy15} use an auxiliary domain as a bridge for transferring between distant domains. However, the data in the auxiliary domain needs to take the form of the co-occurrence data and the learning model is limited to matrix factorization, which greatly limits its application scope. The major differences of the proposed DERWENT method with it are that the DERWENT method does not have the requirement on the data form in auxiliary domains and that the DERWENT method can achieve the multi-step transition in the auxiliary domains while \cite{tszy15} only limits to one step. DDTL \cite{tzpy17} aims to select useful data from the source and auxiliary domains through a selective learning method inspired by self-paced learning to improve the performance in the target domain. The proposed DERWENT model is different from DDTL in mainly two aspects. Firstly, DDTL selects source and auxiliary data based on the idea of self-paced learning but it cannot explicitly give the transfer path between the source and target domains, while the proposed DERWENT method can do that with the help of the deep random walk. Secondly, DDTL selects useful data from the source and auxiliary domains according to the defined reconstruction error, while the proposed DERWENT method relies on the deep random walk with two designed criteria including the similarity between adjacent data in a sequence sampled in the deep random walk and the reconstruction of the ending data point in the sequence based on other data points.

As mentioned in \cite{tzpy17}, Self-Taught Learning (STL) \cite{rblpy07}, which aims to learn a good feature representation through a large amount of unlabeled data, can also work under the distant transfer learning setting where the auxiliary data take the role of unlabeled data. STL is an unsupervised method with the original formulation working with linear sparse coding models. In recent years, with the development of deep learning, STL has adopted deep neural networks as basic models and has achieve better performance as in \cite{kk17,glzl14}. However, distant transfer learning is different from STL in that STL treats the unlabeled data as the source domain, while such unlabeled data is treated as data in the auxiliary domains of distant transfer learning. 

DeepWalk \cite{pas14}  aims to obtain sequences of data nodes in a graph for model training. DeepWalk mainly uses random walk to sample sequences from the graph, and it is to maximize the co-occurrence probability among the data nodes that appear within a window in a sentence in the spirit of the SkipGram model \cite{mccd13}. Different from Deepwalk, the proposed DERWENT method uses random walk for a data graph to generate the transfer path between the source and target domains through auxiliary domains for distant transfer learning and it is to maximize the similarity between adjacent data points in a sampled sequence and to minimize the reconstruction error of the ending data point with respect to other data points in the same sequence via the LSTM.

Tan et al. \cite{tzny14} and Ng et al. \cite{nwy14} use random walk to transfer information between two heterogeneous domains. Those two works are different from ours in two aspects. Firstly, the problem settings are different. Those two works require the existence of co-occurrence data for transferring between two heterogeneous domains, while our work has no such requirement. Secondly, the ways to use random walk are different. Those two works use random walk to compute the probabilities of traversing between source and target instances and then use such probabilities to do the transfer in terms of instances or features, however, our work uses random walk to sample sequences to connect two domains through auxiliary domains and then uses sampled sequences to update the entire network based on three proposed losses.

\section{The DERWENT Model}

In this section, we introduce the proposed DERWENT model.

\subsection{Problem Settings}

By following the problem setting in DDTL \cite{tzpy17} where there are a source domain and a target domain, the source domain has a large labeled training dataset $\mathcal{D}_s=\{(\mathbf{x}^s_i,y^s_i)\}_{i=1}^{n_s}$, where $y^s_i\in\{0,1\}$ is the class label of the $i$th data point $\mathbf{x}^s_i$ in the source domain and $n_s$ denotes the number of data points in the source domain, and the target domain has a small labeled training dataset $\mathcal{D}_t=\{(\mathbf{x}^t_i,y^t_i)\}_{i=1}^{n_t}$, where $y^t_i\in\{0,1\}$ is the class label of the $i$th data point $\mathbf{x}^t_i$ in the target domain and $n_t$ denotes the number of data points in the target domain. Here we assume $n_s\gg n_t$. Since the source and target domains have a large discrepancy, the direct transfer from the source domain to the target domain may have no effect or even negative effect to improve the performance of the target domain. Instead, we assume that there is an auxiliary unlabeled dataset $\mathcal{D}_a=\{\mathbf{x}^a_1,\ldots,\mathbf{x}^a_{n_a}\}$ where $n_a\gg n_t$ denotes the number of data points. This auxiliary dataset contains data points from diverse auxiliary domains and it will act as a bridge to help transfer the knowledge from the source domain to the target domain in order to improve the performance of the target domain.

\subsection{The Model}

To achieve that, we propose the DERWENT model which is based on deep random walk.

In the DERWENT model, we first learn a hidden representation for all the data in the source, auxiliary, and target domains as $\hat{\mathbf{x}}^*_i=\phi(\mathbf{x}^*_i)$, where $\phi(\cdot)$ denotes a feature extraction network, the superscript $*$ in $\mathbf{x}^*_i$ can be $s$, $a$, or $t$.

To measure similarities between data points, we construct a graph $G$ on all the data from all the domains with each data point corresponding to a node in this graph and the edge weight defined as
\begin{equation}
\label{graph_construction}
\begin{aligned}
e(\hat{\mathbf{x}},\hat{\mathbf{x}})=&\ 0\quad \forall \mathbf{x}\in \mathcal{D}_s\cup\mathcal{D}_a\cup\mathcal{D}_t\\
e(\hat{\mathbf{x}}_1,\hat{\mathbf{x}}_2)=&\ \exp\{\mathrm{cos}(\hat{\mathbf{x}}_1,\hat{\mathbf{x}}_2)\}\quad \forall \mathbf{x}_1,\mathbf{x}_2\in \mathcal{D}_s\ \mathrm{or}\ \forall \mathbf{x}_1,\mathbf{x}_2\in \mathcal{D}_t\\
e(\hat{\mathbf{x}}_1,\hat{\mathbf{x}}_2)=&\ \exp\{\eta_1\mathrm{cos}(\hat{\mathbf{x}}_1,\hat{\mathbf{x}}_2)\}\quad \forall \mathbf{x}_1\in \mathcal{D}_s\ \mathbf{x}_2\in\mathcal{D}_a\ \mathrm{or}\ \forall \mathbf{x}_1\in \mathcal{D}_a\ \mathbf{x}_2\in\mathcal{D}_s\\
e(\hat{\mathbf{x}}_1,\hat{\mathbf{x}}_2)=&\ \exp\{\eta_2\mathrm{cos}(\hat{\mathbf{x}}_1,\hat{\mathbf{x}}_2)\}\quad \forall \mathbf{x}_1\in \mathcal{D}_t\ \mathbf{x}_2\in\mathcal{D}_a\ \mathrm{or}\ \forall \mathbf{x}_1\in \mathcal{D}_a\ \mathbf{x}_2\in\mathcal{D}_t,
\end{aligned}
\end{equation}
where $\hat{\mathbf{x}}=\phi(\mathbf{x})$, $\hat{\mathbf{x}}_1=\phi(\mathbf{x}_1)$, $\hat{\mathbf{x}}_2=\phi(\mathbf{x}_2)$, $\eta_1,\eta_2$ are hyperparameters to increase the probability of finding source/target nodes depending on the direction of the random walk as introduced later, and $\mathrm{cos}(\cdot,\cdot)$ denotes the cosine similarity between two vectors, matrices or tensors of the same size with its definition as
$\mathrm{cos}(\mathbf{z}_1,{\mathbf{z}}_2) = \langle{\mathbf{z}}_1,{\mathbf{z}}_2\rangle/(\|\mathbf{z}_1\|\cdot\|\mathbf{z}_2\|)$ where $\langle\cdot,\cdot\rangle$ denotes the dot product.

Based on Eq. (\ref{graph_construction}), there is no edge between the source and target samples and this is because the two domains have a large discrepancy, leading to unreliable similarities. There is no self loop in this graph. Moreover, during the optimization process, $\phi(\cdot)$ changes over epoches and so do edge weights in the graph $G$.

Based on the graph $G$, the random walk works as follows. Suppose we want to traverse the nodes in a graph. At current node $i$, the probability to visit node $j$ next is proportional to $e(i,j)$. So the random walk will start at a node and then randomly visit the next node with such probability until reach some goal node.

In the DERWENT model, we can construct the random walk in two directions. The random walk of the first type starts at a node corresponding to a data point in the source domain, and then randomly visit one of its neighbors with the probability proportional to edge weights. This process will continue until reaching a node in the target domain or the number of nodes visited exceeds a threshold denoted by $\theta$. The random walk of the second type acts similarly but it will start at a node in the target domain and will stop when reaching a node in the source domain or the number of nodes visited exceeds $\theta$.

Here we discuss how to learn from the first type in the DERWENT model. Given a mini-batch which contains a subset of data points from the source, auxiliary, and target domains, we first construct a graph $G$ on this mini-batch with $\eta_1$ being 1 and $\eta_2$ being $\eta$ that is a hyperparameter. Here $\eta>1$ will increase the probability to find a target instance in the neighborhood during the random walk and shorten the length of sequences in the random walk. Then we conduct random walk on the graph $G$ to sample several sequences with different starting nodes. For the $i$th sequence $\mathcal{S}_i=(\hat{\mathbf{x}}_{i,1},\ldots,\hat{\mathbf{x}}_{i,n_{s_i}})$ where $n_{s_i}$ denotes the length of this sequence or equivalently the number of nodes visited satisfying $n_{s_i}\le \theta$, we expect two neighboring data points to be similar, which is to help learn the feature extraction network, and define the corresponding loss function as
\begin{equation}
l_{i,1}=\sum_{j=1}^{n_{s_i}-1}-\ln\sigma_{\alpha}(\mathrm{cos}(\hat{\mathbf{x}}_{i,j},\hat{\mathbf{x}}_{i,j+1}))
-\ln(1-\sigma_{\alpha}(\mathrm{cos}(\hat{\mathbf{x}}_{i,j},\hat{\mathbf{z}}_{i,j}))),\label{loss_similarity}
\end{equation}
where $\sigma_{\alpha}(x)=\frac{1}{1+\exp\{-\alpha x\}}$ denotes a scaled sigmoid function to make the output spread more over $[0,1]$ due to the limited range (i.e., $[-1,1]$) of the cosine similarity, and $\hat{\mathbf{z}}_{i,j}$ is sampled out of $\mathcal{S}_i$ but in the mini-batch to act as a dissimilar data point to $\hat{\mathbf{x}}_{i,j}$.
For $\mathcal{S}_i$, if the last data point is from the target domain, which means that the random walk finds a path from the source domain to the target domain, we expect that this target data point can be represented based on other data points in this sequence since nodes in a sequence generated by deep random walk are inherently related based on the hidden feature representation. To achieve that, we use the sequence of $\mathcal{S}_i$ except $\hat{\mathbf{x}}_{i,n_{s_i}}$ to reconstruct it and hence we can formulate the corresponding loss function as
\begin{equation}
l_{i,2}=\|\hat{\mathbf{x}}_{i,n_{s_i}}-f_d(\mathrm{LSTM}(\hat{\mathbf{x}}_{i,1},\ldots,\hat{\mathbf{x}}_{i,n_{s_i}-1}))\|,\label{loss_LSTM}
\end{equation}
where $\mathrm{LSTM}()$ denotes an LSTM to output the hidden state of the last position and $f_d(\cdot)$ is a neural decoder to generate an approximation of $\hat{\mathbf{x}}_{i,n_{s_i}}$. Here the LSTM is used since $\hat{\mathbf{x}}_{i,1},\ldots,\hat{\mathbf{x}}_{i,n_{s_i}-1}$ form a sequence. Moreover, by defining the set of labeled data in $\mathcal{S}_i$ from either source or target domain as $\mathcal{L}_i$, we formulate the classification loss as
\begin{equation*}
l_{i,3}=\sum_{(\hat{\mathbf{x}},y)\in\mathcal{L}_i}w(\hat{\mathbf{x}})(-y\ln(\sigma(f_c(\hat{\mathbf{x}})))
-(1-y)\ln(1-\sigma(f_c(\hat{\mathbf{x}})))),
\end{equation*}
where $\sigma(x)=\frac{1}{1+\exp\{-x\}}$ denotes the sigmoid function, $f_c(\cdot)$ denotes the classification network and $w(\hat{\mathbf{x}})$, a measure of the instance importance, is equal to 1 for a target data point and otherwise a positive value smaller than 1. For a source data point $\hat{\mathbf{x}}$, we define $w(\hat{\mathbf{x}})$ as $w(\hat{\mathbf{x}})=\sigma_{\alpha}(\mathrm{cos}(\hat{\mathbf{x}},\hat{\mathbf{x}}_{i,n_{s_i}}))$, which reflects the confidence to use the loss of a source data point for the target domain. 

By combining the above three parts, the objective function of the DERWENT model with the first type of the random walk is formulated as
\begin{equation}
\min \sum_{i} l_{i,1}+o_i(\lambda_1 l_{i,2}+\lambda_2l_{i,3}),\label{DERWENT_obj}
\end{equation}
where $\lambda_1,\lambda_2$ are regularization parameters, and the indicator $o_i$ equals 1 when $\mathcal{S}_i$ reaches a target data point and otherwise 0. Parameters to be optimized in problem (\ref{DERWENT_obj}) include those in the feature extraction network $\phi(\cdot)$, the LSTM, the neural decoder $f_d(\cdot)$, and the classification network $f_c(\cdot)$.

The second type of the random walk can be formulated similarly with slight difference. To increase the probability to reach a source node in the graph $G$ and shorten the length of sequences, $\eta_1$ and $\eta_2$ are set to $\eta$ and 1, respectively. $l_{i,1}$ and $l_{i,3}$ have no change. $l_{i,2}$ is formulated similarly with the ending data point $\hat{\mathbf{x}}_{i,n_{s_i}}$ from the source domain being represented by other data points in the same sequence. In the combined objective function, the starting target data point will contribute the classification loss no matter whether the sequence reaches the source domain.

\begin{figure}[!htb]
    \centering
    \includegraphics[scale=0.45]{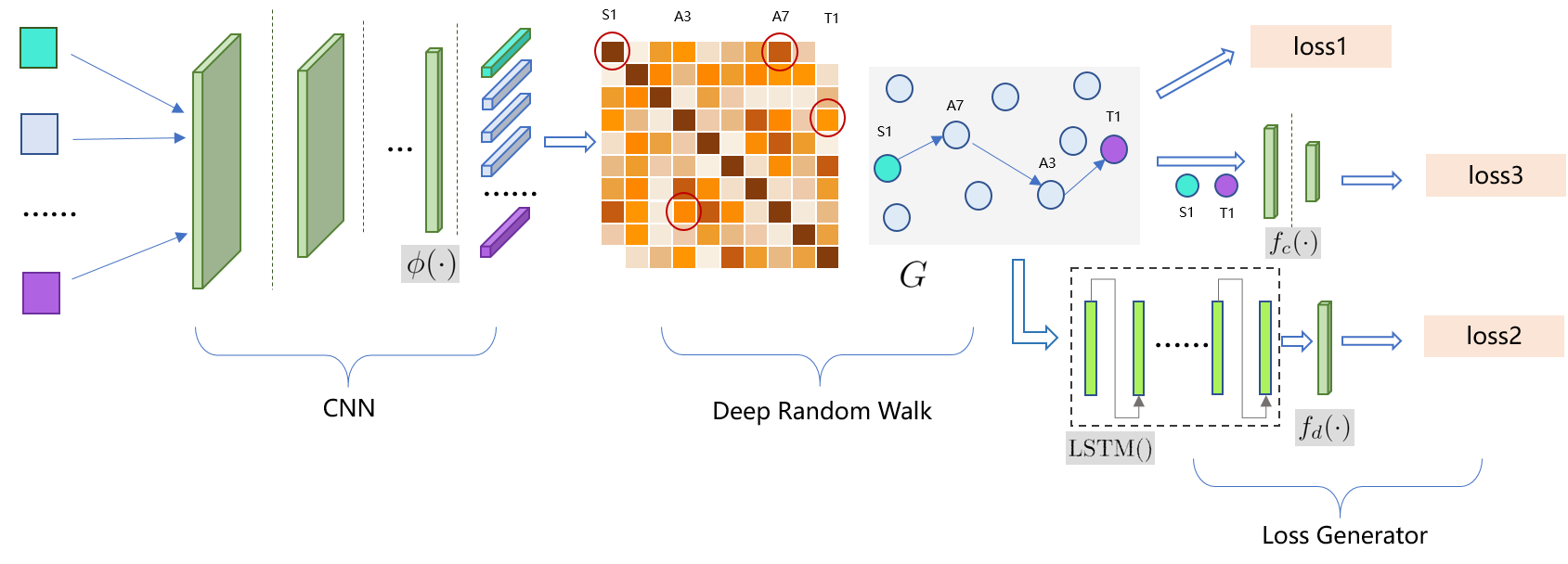}
    \vskip -0.1in
    \caption{The architecture of the DERWENT model consisting of three parts. (1) Images features are extracted by a pre-trained deep convolutional neural network followed by the feature extraction network $\phi(\cdot)$. (2) According to the hidden feature representation generated in the previous step, we construct the data graph $G$ for each mini-batch and conduct the random walk on $G$ to generate sequence. (3) Three losses are calculated by the loss generator component on the resulting sequences.}
    \label{fig:label2}
    \vskip -0.1in
\end{figure}

The entire objective function of the DERWENT model is to sum those of the two types together. In summary, the architecture of the DERWENT model is illustrated in Figure \ref{fig:label2}.

\subsection{Discussion}

Different from DDTL, the DERWENT model identifies the transfer path $\{\mathcal{S}_i\}$ between the source and target domains in two directions based on the random walk technique, learns good feature extraction network via two losses $l_{i,1}$ and $l_{i,2}$, and reuses the label information in the source domain by defining the weight function $w(\cdot)$ based on the learned feature extraction network. Different from the DeepWalk method which maximizes the co-occurrence probability among the data nodes, the DERWENT method maximizes the similarity between adjacent data points in a sampled sequence via loss $l_{i,1}$ and minimizes the reconstruction error of the ending data point via loss $l_{i,2}$.

\section{Experiments}



We conduct experiments on three benckmark datasets, including the Animals with Attributes (AwA) dataset \cite{lnh09}, the Caltech-256 dataset \cite{ghp07} and the CIFAR-100 dataset \cite{krizhevsky09}.
The AwA dataset contains 30,475 pictures with 50 categories, where the number of instances per class varies from 92 to 1,168. 
We select one of three categories including `chihuahua’, `sheep’ and `lion’ to form the positive class of the source domain, and select one of six categories including `antelope’, `chimpanzee’, `rabbit’, `bobcat’, `pig’ and `german+shepherd’ as the positive class of the target domains. Moreover, by following \cite{tzpy17}, we mix data from seven categories `beaver’, `blue+whale’, `mole’, `mouse’, `ox’, `skunk’ and `weasel’ to form the negative class for source and target domains but with no overlapping. Data of all the remaining categories are used as auxiliary domains.
The Caltech-256 dataset contains 30,607 images with 257 categories, including a background category `clutter'. There are 80 to 827 images in each category. To validate the performance between distant domains, we select some relatively different categories to form the source and target domains, such as `baseball-bat', `conch', `airplane’, `skateboard’, `soccer-ball’, `horse’ and `gorilla’. Specifically, we first randomly select a category as the positive class of source domain and then randomly select another category to be the positive class of target domain. Data in the `clutter' category are randomly selected to form negative instances for both source and target domains but with no overlapping. Data of all the remaining categories are used as auxiliary domains.
The CIFAR-100 dataset contains 100 classes, where the number of instances per class is 500. We select one category from `chair', `bus', `rose', `woman' and `bottle' to form the positive class of source domain and select one of three categories including `cup', `phone' and `bowl' as the positive class of target domain. Then we randomly choose one category as the positive example of the source domain and the target domain respectively. We mix data from categories related to aquatic mammals including `beaver', `dolphin', `otter', `seal' and `whale' to form negative examples for source and target domains with no overlapping. Data of all the remaining categories are used as auxiliary domains.
According to the above construction of different domains, on the AwA dataset, we have 9 distant transfer learning tasks, including `chihuahua-to-bobcat' (C$\rightarrow$B), `chihuahua-to-antelope' (C$\rightarrow$A), `chihuahua-to-pig' (C$\rightarrow$P), `sheep-to-rabbit' (S$\rightarrow$R), `sheep-to-chimpanzee' (S$\rightarrow$CH), `sheep-to-german+shepherd' (S$\rightarrow$SH), ‘lion-to-rabbit' (L$\rightarrow$R), `lion-to-chimpanzee' (L$\rightarrow$CH) and `lion-to-german+shepherd' (L$\rightarrow$SH). On Caltech-256 dataset, we have 6 distant transfer learning tasks, including `airplane-to-soccer-ball' (A$\rightarrow$S), `gorilla-to-baseball-bat' (G$\rightarrow$B), `airplane-to-skateboard' (A$\rightarrow$SK), `horse-to-conch' (H$\rightarrow$C), `soccer-ball-to-skateboard' (S$\rightarrow$SK) and `soccer-ball-to-conch' (S$\rightarrow$C). On the CIFAR-100 dataset, we have 5 distant transfer learning tasks, including `bus-to-phone' (B$\rightarrow$P), `chair-to-cup' (C$\rightarrow$CU), `rose-to-phone' (R$\rightarrow$P), `bottle-to-bowl' (BT$\rightarrow$BW) and `woman-to-phone' (W$\rightarrow$P) .

The baseline models in comparison include a deep neural network (DNN) which is trained on the target data only, DAN \cite{lcwj15}, DANN \cite{gyue16}, CNN-based STL \cite{kk17}, and DDTL. We also compare with a varaint of the proposed DERWENT method by discarding the second loss defined in Eq. (\ref{loss_LSTM}) and we denote it by \emph{DERWENT w/o LSTM}.  
We use the VGG-11 model \cite{sz14} pre-trained on the ImgeNet dataset before the feature extraction network $\phi(\cdot)$, which has a Fully-Connected (FC) layer with 256 hidden units and the activation function as the tanh function. We use the same network structure for all the baseline models. In the DERWENT model, we use a one-layer bi-directional LSTM with 128 hidden units which is used to compute the second loss defined in Eq. (\ref{loss_LSTM}), the neural decoder $f_d(\cdot)$ has a FC layer with 256 outputs, and the classification network $f_c(\cdot)$ has a FC layer with 2 outputs.  
For optimization, we use the mini-batch SGD with the Nestorov momentum 0.9. The batch size is set to 128, including 10, 8 and 110 in source, target and auxiliary domain respectively. The learning rate of the classifiers are set 10 times to that of the feature extractor by following \cite{gyue16}. $\eta$ in the graph (i.e., Eq. (\ref{graph_construction})) is initialized to 1.1 and then increased according to epochs as  ${1.1}^{\lfloor \mathrm{epochs}/3 \rfloor}$. All the regularization parameters in the DERWENT model is set to 1.



In each experiment, we randomly selected 10 labeled instances of each class in the target domain for training and the rest for testing. Each setting is repeated for three times and the average results are reported in Tables \ref{AWA2-table}-\ref{Cifar-table}. According to the results, we can see that the accuracy of DAN and DANN that are transfer learning methods is lower than that of DNN, resulting in `negative transfer'. This is because that there is a large discrepancy between the source and target domains. The STL method performs slightly better than DNN as it can learn a useful feature representation from auxiliary domains.  
As a distant transfer learning method, DDTL performs better than DNN, DAN, DANN and STL as it uses auxiliary domains as a bridge to help transfer the knowledge contained in the source domain to help the learning in the target domain. Among all the methods in comparison, the proposed DERWENT method performs the best,\footnote{For clear presentation, we do not include standard deviations in Tables \ref{AWA2-table}-\ref{Cifar-table}. We have conducted the $t$ test to verify that DERWENT is significantly better than baseline models.} which demonstrates the effectiveness of the proposed model. In an ablation study to test the effectiveness of the second loss (i.e., Eq. (\ref{loss_LSTM})) based on LSTM, we can see that DERWENT performs better than that without the second loss (corresponding to the DERWENT w/o LSTM method) in most settings and hence the second loss is useful to improve the performance.  

\begin{figure}[!htbp]
    \centering
    \includegraphics[scale=0.435]{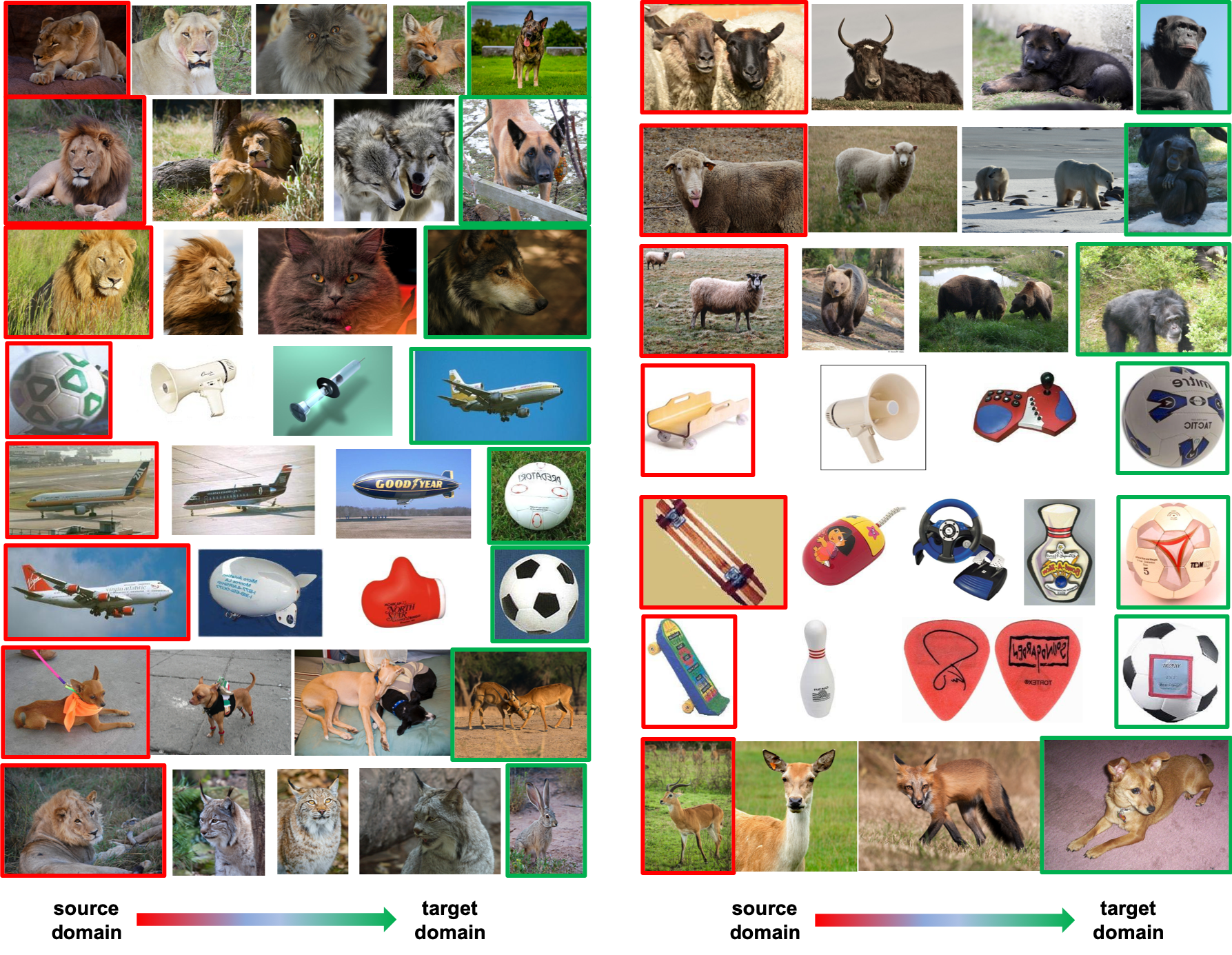}
    \caption{Selected sequences generated by the DERWENT method. Specifically, each row represents a transfer sequence from the source domain in a red rectangle to the target domain in a green rectangle.}
    \label{fig:samples}
\end{figure}

In order to understand how to transfer between distant domains through auxiliary domains, we visualize in Figure \ref{fig:samples} the transfer sequences obtained by the random walk in the DERWENT method. 
According to Figure \ref{fig:samples}, we can see that in each sequence, the source image in a red rectangle is completely different from the target image in a green rectangle, 
and from left to right, the images visited by the random walk are gradually close to the target image. For example, in two tasks `airplane-to-soccer-ball' (located in the third row from the bottom in the left part of Figure \ref{fig:samples}) and `skateboard-to-soccer-ball' (located in the second row from the bottom in the right part of Figure \ref{fig:samples}), the source and target domains are quite different. The DERWENT method first relates source images to the `blimp' and `bowling-pin' classes, respectively, which are similar to source images, and then gradually visits images that are more similar to the target domain until reaching some target images.

\begin{figure}[!htbp]
    \centering
    \includegraphics[scale=0.45]{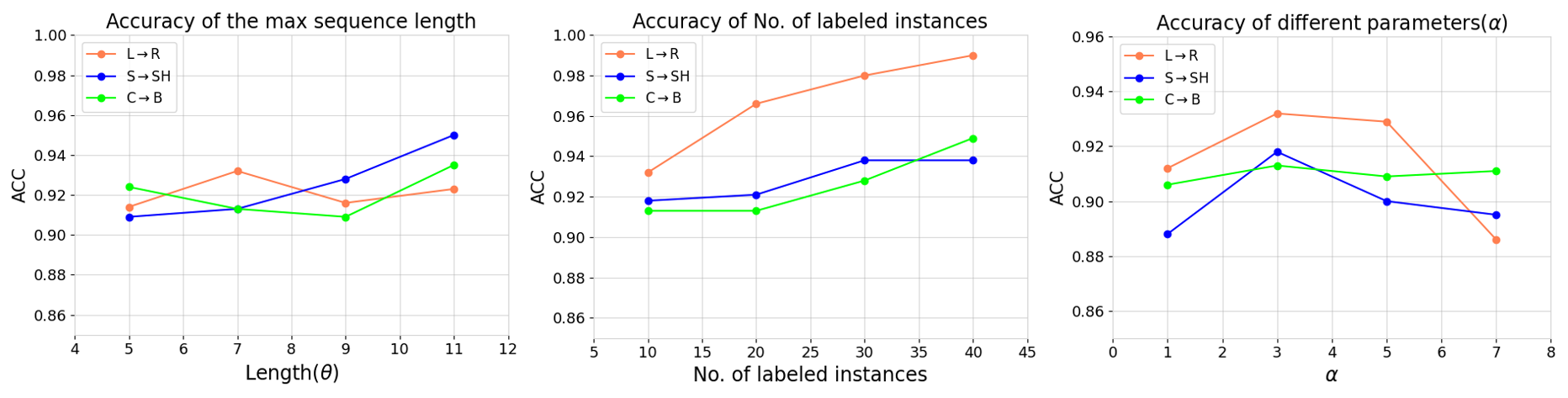}
    \caption{Sensitivity analysis of hyperparameters of the DERWENT algorithm.}
    \label{fig:acc_params}
\end{figure}

To test the sensitivity of the transfer performance of the DERWENT model with respect to different hyperparameters including the maximum length $\theta$ of sampled sequences in the random walk, number of labeled instances in the target domain, and $\alpha$ used in the first loss (i.e., Eq. (\ref{loss_similarity})), we conduct experiments for each hyperparameter by fixing other hyperparameters on three distant transfer learning tasks, including L$\rightarrow$R, S$\rightarrow$SH and C$\rightarrow$B. According to the results shown in Figure \ref{fig:acc_params}, we can see that $\theta$ has little effect on the performance and one possible reason is that the random walk has reached the destination in less than $\theta$ steps. It is easy to understand that more labeled instances in the target domain lead to better performance. Moreover, according to the results, setting $\alpha$ to 3 has the best performance for the three tasks and this is the setting for $\alpha$ in all the experiments.

\begin{table}[!htbp]
\caption{Accuracy (\%) of different models on different tasks of the AwA dataset.}
\label{AWA2-table}
\centering
\resizebox{\textwidth}{!}{
\begin{tabular}{@{}lllllllllll@{}}
\toprule
Method & C$\rightarrow$B & C$\rightarrow$A & C$\rightarrow$P & S$\rightarrow$R & S$\rightarrow$CH & S$\rightarrow$SH & L$\rightarrow$R & L$\rightarrow$CH & L$\rightarrow$SH & Avg \\ \midrule
DNN&  83.5&  89.1&  65.0&  87.3&  76.0&  77.5&  87.3&  76.0&  77.5& 79.9   \\
DAN&  57.9& 75.9& 47.8& 82.5& 54.6& 68.6&  71.5&  67.1&  70.7& 66.3  \\
DANN&  60.3& 67.0& 68.4& 48.5& 60.0& 55.7& 45.1&  55.8&  64.8& 58.4  \\
STL&  83.1&  89.8& 64.4&  89.1&  79.9&  80.5&  89.1&  79.9&  80.5& 81.8  \\
DDTL&  85.6& 92.9& 77.2& 89.3& 72.5& 79.2& 91.8& 78.7& 80.8& 83.1  \\
\textbf{DERWENT  w/o LSTM}&  \textbf{91.3}&  96.1&  75.9&  94.2&  85.2&  \textbf{91.9}&  93.2&  88.7&  \textbf{92.7}& 89.9   \\
\textbf{DERWENT}&  90.3&  \textbf{96.3}&  \textbf{77.9}&  \textbf{94.6}&  \textbf{92.7}&  91.8&  \textbf{95.2}&  \textbf{89.4}&  92.0& \textbf{91.1}  \\
\bottomrule
\end{tabular}
}
\end{table}

\begin{table}[!htbp]\small
\caption{Accuracy (\%) of different models on different tasks of the Caltech-256 dataset.}
\label{Catech-table}
\centering
\begin{tabular}{@{}llllllll@{}}
\toprule
Method & A$\rightarrow$S & G$\rightarrow$B & A$\rightarrow$SK & H$\rightarrow$C & S$\rightarrow$SK & S$\rightarrow$C & Avg  \\ \midrule
DNN&  82.9&   72.6&   66.7&    82.8&   66.7&   82.8& 75.8    \\
DAN&  81.7& 58.1&  78.5& 75.2& 72.0& 76.3& 73.6    \\
DANN&  49.7&  63.3&  77.4&  82.4& 74.6&  81.0&  71.4   \\
STL&  84.1&   76.1&   69.9&    75.3&   69.9&    75.3& 75.1      \\
DDTL&  84.1& 71.8& 78.5& 89.2& 61.3& 84.9& 78.3    \\
\textbf{DERWENT w/o LSTM}&  \textbf{90.8}&     80.3&     81.7&      \textbf{90.3}&     \textbf{77.4}&      89.2& 84.1         \\
\textbf{DERWENT}&  \textbf{90.8}&     \textbf{85.4}&    \textbf{84.9}&      87.1&     \textbf{77.4}&      \textbf{91.4}& \textbf{86.2}        \\
\bottomrule
\end{tabular}
\end{table}

\begin{table}[!htbp]\small
\caption{Accuracy (\%) of different models on different tasks of the CIFAR-100 dataset.}
\label{Cifar-table}
\centering
\begin{tabular}{@{}lllllll@{}}
\toprule
Method & B$\rightarrow$P & C$\rightarrow$CU & R$\rightarrow$P & BT$\rightarrow$BW & W$\rightarrow$P & Avg  \\ \midrule
DNN&  89.7&  87.4&  89.7&   87.2&   89.7&   88.7    \\
DAN&  85.8&  63.1&  68.2&    79.8&   75.9&   74.6  \\
DANN&  90.5& 88.2& 81.3& 88.7& 87.5& 87.2 \\
STL&  89.2&  87.8&   89.2&    86.2&   89.2&   88.3       \\
DDTL&  91.5&  86.0&  88.0&  83.3&  94.4&  88.6       \\
\textbf{DERWENT w/o LSTM}&  93.4&    \textbf{91.3}&     92.2&      90.5&     96.3&   92.7          \\
\textbf{DERWENT}&  \textbf{93.8}&    91.1&    \textbf{93.0}&      \textbf{91.5}&     \textbf{96.9}&  \textbf{93.3 }           \\
\bottomrule
\end{tabular}
\end{table}

\section{Conclusion}

To solve the distant transfer learning problem, we propose a DERWENT method based on deep random walk, which can help transfer knowledge from the source domain to the target domain across auxiliary domains gradually. Different from existing methods, the proposed DERWENT method can automatically find the transfer path. 
The proposed DERWENT method has shows state-of-the-art performance in three benchmark image datasets. In the future research, we are interested in extending the DERWENT model to handle more general cases with multiple source domains.

\bibliographystyle{plain}
\bibliography{DERWENT}

\begin{thebibliography}{10}

\bibitem{dyxy07}
Wenyuan Dai, Qiang Yang, Gui{-}Rong Xue, and Yong Yu.
\newblock Boosting for transfer learning.
\newblock In {\em Proceedings of the Twenty-Fourth International Conference on
  Machine Learning}, pages 193--200, 2007.

\bibitem{glzl14}
Jun{-}Ying Gan, Lichen Li, Yikui Zhai, and Yinhua Liu.
\newblock Deep self-taught learning for facial beauty prediction.
\newblock {\em Neurocomputing}, 144:295--303, 2014.

\bibitem{gyue16}
Yaroslav Ganin, Evgeniya Ustinova, Hana Ajakan, Pascal Germain, Hugo
  Larochelle, Fran{\c{c}}ois Laviolette, Mario Marchand, and Victor~S.
  Lempitsky.
\newblock Domain-adversarial training of neural networks.
\newblock {\em J. Mach. Learn. Res.}, 17:59:1--59:35, 2016.

\bibitem{ghp07}
Gregory Griffin, Alex Holub, and Pietro Perona.
\newblock Caltech-256 object category dataset.
\newblock Technical Report CNS-TR-2007-001, California Institute of Technology,
  2007.

\bibitem{hs97}
Sepp Hochreiter and J{\"{u}}rgen Schmidhuber.
\newblock Long short-term memory.
\newblock {\em Neural Computation}, 9(8):1735--1780, 1997.

\bibitem{hy11}
Derek~Hao Hu and Qiang Yang.
\newblock Transfer learning for activity recognition via sensor mapping.
\newblock In {\em {IJCAI} 2011, Proceedings of the 22nd International Joint
  Conference on Artificial Intelligence}, pages 1962--1967, 2011.

\bibitem{kk17}
Ronald Kemker and Christopher Kanan.
\newblock Self-taught feature learning for hyperspectral image classification.
\newblock {\em {IEEE} Trans. Geosci. Remote. Sens.}, 55(5):2693--2705, 2017.

\bibitem{kh16}
Mohammad Nazmul~Alam Khan and Douglas~R. Heisterkamp.
\newblock Adapting instance weights for unsupervised domain adaptation using
  quadratic mutual information and subspace learning.
\newblock In {\em 23rd International Conference on Pattern Recognition}, pages
  1560--1565, 2016.

\bibitem{krizhevsky09}
Alex Krizhevsky.
\newblock Learning multiple layers of features from tiny images.
\newblock Technical report, 2009.

\bibitem{kpk10}
M.~Pawan Kumar, Benjamin Packer, and Daphne Koller.
\newblock Self-paced learning for latent variable models.
\newblock In {\em Advances in Neural Information Processing Systems 23}, pages
  1189--1197, 2010.

\bibitem{lnh09}
Christoph~H. Lampert, Hannes Nickisch, and Stefan Harmeling.
\newblock Learning to detect unseen object classes by between-class attribute
  transfer.
\newblock In {\em Proceedings of {IEEE} Computer Society Conference on Computer
  Vision and Pattern Recognition}, pages 951--958, 2009.

\bibitem{lcwj15}
Mingsheng Long, Yue Cao, Jianmin Wang, and Michael~I. Jordan.
\newblock Learning transferable features with deep adaptation networks.
\newblock In {\em Proceedings of the 32nd International Conference on Machine
  Learning}, pages 97--105, 2015.

\bibitem{lwcsy16}
Mingsheng Long, Jianmin Wang, Yue Cao, Jia{-}Guang Sun, and Philip~S. Yu.
\newblock Deep learning of transferable representation for scalable domain
  adaptation.
\newblock {\em {IEEE} Trans. Knowl. Data Eng.}, 28(8):2027--2040, 2016.

\bibitem{mccd13}
Tomas Mikolov, Kai Chen, Greg Corrado, and Jeffrey Dean.
\newblock Efficient estimation of word representations in vector space.
\newblock In {\em Workshop Track Proceedings of the 1st International
  Conference on Learning Representations}, 2013.

\bibitem{nwy14}
Michael~K. Ng, Qingyao Wu, and Yunming Ye.
\newblock Co-transfer learning via joint transition probability graph based
  method.
\newblock In {\em In Proceedings of the 1st International Workshop on Cross
  Domain Knowledge Discovery in Web and Social Network Mining}, pages 1--9,
  2012.

\bibitem{psyt08}
Sinno~Jialin Pan, Dou Shen, Qiang Yang, and James~T. Kwok.
\newblock Transferring localization models across space.
\newblock In {\em Proceedings of the Twenty-Third {AAAI} Conference on
  Artificial Intelligence}, pages 1383--1388, 2008.

\bibitem{pwty09}
Sinno~Jialin Pan, Ivor~W. Tsang, James~T. Kwok, and Qiang Yang.
\newblock Domain adaptation via transfer component analysis.
\newblock In {\em Proceedings of the 21st International Joint Conference on
  Artificial Intelligence}, pages 1187--1192, 2009.

\bibitem{py10}
Sinno~Jialin Pan and Qiang Yang.
\newblock A survey on transfer learning.
\newblock {\em IEEE Transactions on Knowledge and Data Engineering},
  22(10):1345--1359, 2010.

\bibitem{plwy11}
Weike Pan, Nathan~Nan Liu, Evan~Wei Xiang, and Qiang Yang.
\newblock Transfer learning to predict missing ratings via heterogeneous user
  feedbacks.
\newblock In {\em Proceedings of the 22nd International Joint Conference on
  Artificial Intelligence}, pages 2318--2323, 2011.

\bibitem{pas14}
Bryan Perozzi, Rami Al{-}Rfou, and Steven Skiena.
\newblock Deepwalk: Online learning of social representations.
\newblock In {\em Proceedings of The 20th {ACM} {SIGKDD} International
  Conference on Knowledge Discovery and Data Mining}, pages 701--710, 2014.

\bibitem{rblpy07}
Rajat Raina, Alexis Battle, Honglak Lee, Benjamin Packer, and Andrew~Y. Ng.
\newblock Self-taught learning: transfer learning from unlabeled data.
\newblock In {\em Proceedings of the Twenty-Fourth International Conference on
  Machine Learning}, pages 759--766, 2007.

\bibitem{sz14}
Karen Simonyan and Andrew Zisserman.
\newblock Very deep convolutional networks for large-scale image recognition.
\newblock In {\em Proceedings of the 3rd International Conference on Learning
  Representations}, 2015.

\bibitem{tszy15}
Ben Tan, Yangqiu Song, Erheng Zhong, and Qiang Yang.
\newblock Transitive transfer learning.
\newblock In {\em Proceedings of the 21th {ACM} {SIGKDD} International
  Conference on Knowledge Discovery and Data Mining}, pages 1155--1164, 2015.

\bibitem{tzpy17}
Ben Tan, Yu~Zhang, Sinno~Jialin Pan, and Qiang Yang.
\newblock Distant domain transfer learning.
\newblock In {\em Proceedings of the Thirty-First {AAAI} Conference on
  Artificial Intelligence}, pages 2604--2610, 2017.

\bibitem{tzny14}
Ben Tan, Erheng Zhong, Michael~K. Ng, and Qiang Yang.
\newblock Mixed-transfer: Transfer learning over mixed graphs.
\newblock In {\em Proceedings of the 2014 {SIAM} International Conference on
  Data Mining}, pages 208--216, 2014.

\bibitem{thds15}
Eric Tzeng, Judy Hoffman, Trevor Darrell, and Kate Saenko.
\newblock Simultaneous deep transfer across domains and tasks.
\newblock In {\em Proceedings of {IEEE} International Conference on Computer
  Vision}, pages 4068--4076, 2015.

\bibitem{ud10}
Diego Uribe.
\newblock Domain adaptation in sentiment classification.
\newblock In {\em The Ninth International Conference on Machine Learning and
  Applications}, pages 857--860, 2010.

\bibitem{yzdp20}
Qiang Yang, Yu~Zhang, Wenyuan Dai, and Sinno~Jialin Pan.
\newblock {\em Transfer Learning}.
\newblock Cambridge University Press, 2020.

\bibitem{zllj19}
Yuchen Zhang, Tianle Liu, Mingsheng Long, and Michael~I. Jordan.
\newblock Bridging theory and algorithm for domain adaptation.
\newblock In {\em Proceedings of the 36th International Conference on Machine
  Learning}, pages 7404--7413, 2019.

\end{thebibliography}





\end{document}